\documentclass[letterpaper, 10pt, conference]{ieeeconf}
\IEEEoverridecommandlockouts
\usepackage{cite}
\usepackage{amsmath,amssymb,amsfonts}
\usepackage{graphicx}
\usepackage[table,xcdraw]{xcolor}
\usepackage{textcomp}
\usepackage{xcolor}
\usepackage[export]{adjustbox}
\usepackage{multirow}
\usepackage{flushend}
\usepackage{algorithm}
\usepackage{balance}
\usepackage{url}
\usepackage[noend]{algpseudocode}
\usepackage{booktabs}
\newcommand{\removeFigSpace}{\vspace{-0.3cm}}

\begin{document}

\title{\LARGE \bf
Preliminary Evaluation of an Ultrasound-Guided Robotic System for \\Autonomous Percutaneous Intervention
}

\author{Pratima Mohan, Aayush Agrawal and Niravkumar A. Patel
\textit{Member IEEE}
\thanks{*This work was funded by the Science and Engineering Research Board, Department of Science and Technology, Government of India, grant SRG/2021/002086.}
\thanks{Authors are with the Indian Institute of Technology Madras, Chennai, TN 600036, India
{\tt\small [nirav.robotics@gmail.com]}}
}


\maketitle

\begin{abstract} 
Cancer cases have been rising globally, resulting in nearly 10 million deaths in 2023. Biopsy, crucial for diagnosis, is often performed under ultrasound (US) guidance, demanding precise hand coordination and cognitive decision-making. Robot-assisted interventions have shown improved accuracy in lesion targeting by addressing challenges such as noisy 2D images and maintaining consistent probe-to-surface contact. Recent research has focused on fully autonomous robotic US systems to enable standardized diagnostic procedures and reproducible US-guided therapy. This study presents a fully autonomous system for US-guided needle placement capable of performing end-to-end clinical workflow. The system autonomously: 1) identifies the liver region on the patient’s abdomen surface, 2) plans and executes the US scanning path using impedance control, 3) localizes lesions from the US images in real-time, and 4) targets the identified lesions, all without human intervention. This study evaluates both position and impedance-controlled systems. Validation on agar phantoms demonstrated a targeting error of $5.741 \pm 2.70 $ mm, highlighting its potential for accurately targeting tumors larger than 5 mm. Achieved results show its potential for a fully autonomous system for US-guided biopsies.

\end{abstract}

\vspace{2pt}
\textbf{\textit{Keywords- Autonomous, percutaneous interventions, ultrasound-guided robot, biopsy}}

\section{INTRODUCTION}

Biopsy, a common method for cancer diagnosis, involves needle insertion to extract tissue samples for pathological examination. It is often guided by ultrasound (US) due to its cost-effectiveness, portability, and lack of radiation. The clinician holds the US probe in one hand while inserting the needle with the other, interpreting real-time US images based on the understanding of general human anatomy. Maintaining needle alignment within the imaging plane while ensuring accuracy is difficult, especially for less experienced surgeons, leading to potential errors, prolonged procedures, and risks like internal bleeding. The procedure's reliance on operator skill also increases the risk of fatigue and musculoskeletal injuries \cite{graumann2016robotic, chen2021ultrasound}. Accurate access to small targets is essential to avoid critical structures like blood vessels and nerve plexuses.

\begin{figure}[htbp!]
    \centering
    \includegraphics[width=\columnwidth]{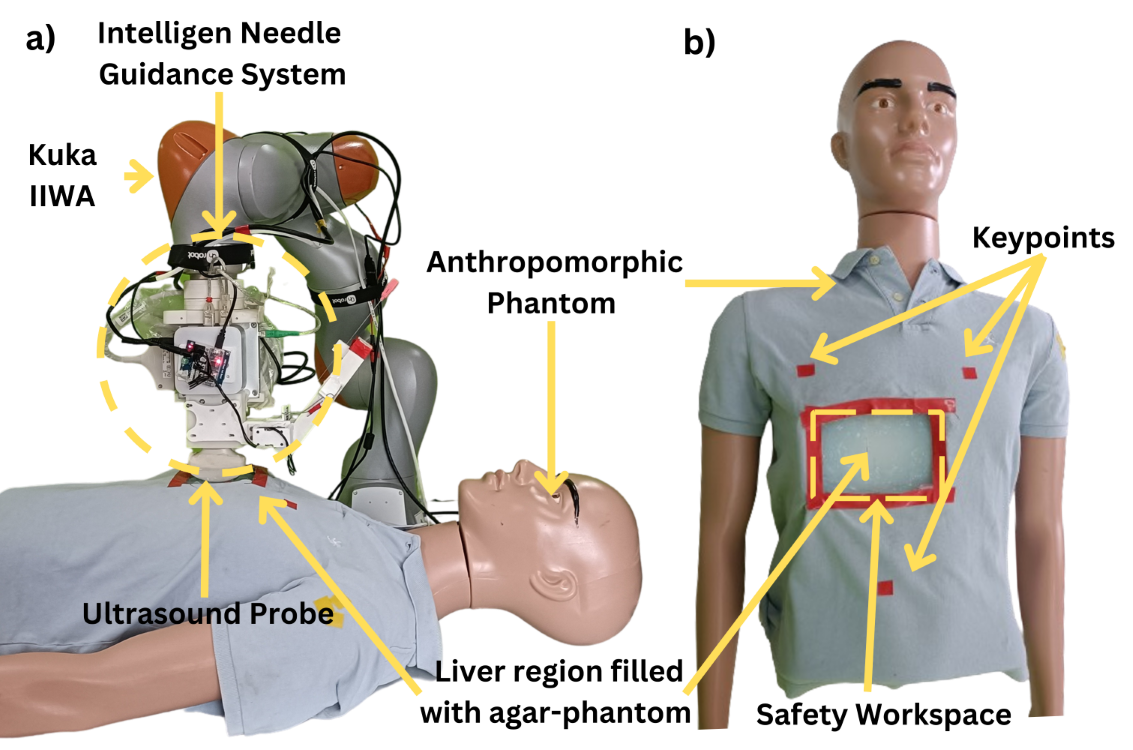}
    \caption{Ultrasound-guided robotic system for autonomous percutaneous interventions showing: (a) a KUKA iiwa7 with Intelligent Needle Guidance system, an ultrasound probe and mannequin setup for the experiments, and (b) top view of the mannequin showing key points for organ localization and the desired safe scan region for the liver interventions.}
    \removeFigSpace
    \label{fig:exp_setup_manqn}
\end{figure}

\begin{figure*}[ht!]
    \centering
    \includegraphics[width=\textwidth]{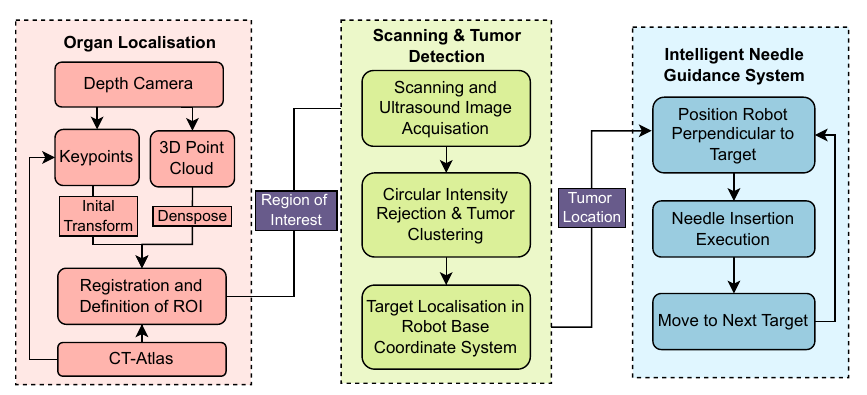}
    \caption{System architecture for US-guided autonomous percutaneous intervention showing its modules for: (1) organ localization, (2) trajectory panning and lesion detection, and (3) needle manipulation.}
    \removeFigSpace
    \label{fig:sys_arch_block_dia}
\end{figure*}

Robot-assisted minimally invasive interventions \cite{chen2020deep, virga2016automatic, chen2021ultrasound} have been shown to outperform human capabilities by providing enhanced dexterity and precision, leading to more accurate procedures and improved outcomes. Over the past decade, research has focused on robotic US screening to improve accuracy and reduce hand-eye coordination challenges in manual percutaneous procedures \cite{huang2018robotic, mustafa2013development, boctor2008three}. However, robot-assisted percutaneous needle placement faces significant challenges including low signal-to-noise ratio of ultrasound images, echo artifacts, and difficulties in maintaining proper probe-to-surface contact. Optimal US image quality is achieved when the probe is positioned perpendicular to the contact surface \cite{jiang2021deformation, jiang2021motion}. Traditional 2D ultrasound (US) images lack the spatial information needed for clinicians to fully visualize the 3D anatomy, making it cognitively challenging to localize lesions. This limitation has led to the adoption of 3D US technology, which can be achieved either through a specialized 3D probe with a 2D transducer array or by capturing multiple 2D images with a standard probe and combining them into a 3D volume. This process uses tracking systems like optical, electromagnetic, or mechanical trackers to combine pose information with each 2D image, allowing accurate alignment and reconstruction into a 3D volume. The quality of 3D volumes generated by a 3D US probe is comparable to those created by compounding tracked B-mode images from a 2D probe. These factors motivated the use of tracked robotic 3D US built from 2D images instead of commercial 3D US probes \cite{boctor2008three}.

Recent research has increasingly focused on addressing the challenges of ultrasound (US) scanning by shifting from robot-assisted to fully autonomous robotic systems. This transition aims to reduce or eliminate human involvement, improving both consistency and accuracy while minimizing dependence on operator skill and experience. US imaging remains highly user-dependent and lacks reproducibility, making it difficult to standardize diagnostic procedures. This issue is further complicated by anatomical variability, uneven body surface geometry, and skin elasticity, all of which affect the quality and reliability of US imaging \cite{suligoj2021robust}. These factors also hinder the automation of volumetric data analysis, which is crucial for integrating with global diagnostic systems and computer-assisted interventions.

Mustafa et al. developed an algorithm for autonomous liver scanning by detecting the umbilicus and estimating the epigastric region \cite{mustafa2013development}. Huang et al. developed an RGB-based color segmentation technique combined with a zig-zag ultrasound scanning approach \cite{huang2018fully, huang2018robotic}. Virga et al. demonstrated a fully autonomous US scanning framework with MRI-based atlas registration and force-controlled scanning, validating on human volunteers \cite{virga2016automatic}. Tan et al. developed an autonomous breast scanning system using statistical modeling for chest localization, but its complexity and reliance on 3D path computations limit clinical viability \cite{tan2022flexible}. Suligoj et al. advanced US automation with multi-scan line generation, impedance control, and automatic data analysis, but it still requires manual region selection in MRI/CT volumes, hindering full autonomy \cite{suligoj2021robust}. Kojcev's dual-robot system addresses autonomous needle placement but requires a complex calibration setup and a large operating room \cite{kojcev2016dual}. Chen et al. developed a portable puncture robot that improves accuracy by converting manually planned ultrasound paths into robotic coordinates, validated on ex-vivo tissues \cite{chen2021ultrasound}. However, pre-interventional assessment and post-interventional decision-making still lack the level of autonomy needed for practical clinical application \cite{tan2022flexible, ma2021autonomous}. Ongoing research into autonomous US-guided needle placement aims to achieve reproducible diagnostic outcomes with clinical applicability. Li et al. and Haxthausen et al. have classified US-guided robotic systems into levels of robot autonomy (LORA), ranging from levels 1-2 for assisted teleoperation to levels 3-4 for collaborative manipulation by robot and clinician, and levels 5-9 for assisted autonomy based on clinician involvement \cite{von2021medical ,li2021overview}. The highest LORA for ultrasound therapy systems is level 7 \cite{kojcev2016dual}. However, no US-guided robotic system for percutaneous intervention has yet validated fully autonomous end-to-end needle placement without any human involvement.

This paper presents a fully autonomous system for ultrasound-guided robotic percutaneous interventions, enabling needle placement without human involvement. To the best of our knowledge, this is the first time that a fully autonomous US-guided robotic system for needle placement has been demonstrated. The system uses a depth camera to create a point cloud of the patient, which is registered to a CT-based atlas to localize the anatomical region of interest (ROI). The robot autonomously plans and executes a multi-sweep trajectory for scanning the ROI, maintaining continuous probe-to-surface contact using impedance control. During the scan, US images are generated and processed in real time to detect the lesions using CIR (Circular Intensity Rejection) algorithm. All the images are stacked to produce 3D volume, which is processed using a clustering algorithm to identify lesions, calculating the centroid of each target. The robot then targets the centroid of each detected lesion, and finally, the needle guidance module inserts the needle precisely into the target tissue for sampling. 

The key contributions of the paper are as follows:
\begin{enumerate}
\item A fully autonomous robotic system for US-guided percutaneous interventions is demonstrated in anthropomorphic phantom studies as shown in Fig.  \ref{fig:exp_setup_manqn}.
\item ATIR (Atlas-based Transformation and Iterative Registration)A fully autonomous multi-organ localization has been demonstrated by registering real-time point cloud and MRI/CT. 
\item Autonomous robot trajectory control for US-imaging to produce a 3D volumetric representation of the identified organ and lesion segmentation has been achieved.  
\item End-to-end clinical workflow is demonstrated with fully autonomous needle placement in anthropomorphic phantom studies. 
\end{enumerate}

\section{Material and Methods}
\label{sec:m&m}

This section gives an overview of the proposed robotic system for autonomous liver biopsy procedures. Figure \ref{fig:sys_arch_block_dia} shows the system framework and its modules. The rest of the section is organized as follows: (A) Hardware Setup (B) ATIR: Atlas-based Transformation and Iterative Registration (C) Robot path planning and execution (D) Lesion segmentation and 3D localization, (E) INGS: Intelligent Needle Guidance System.

\begin{figure}[t!]
    \centering
    \includegraphics[width=\columnwidth]{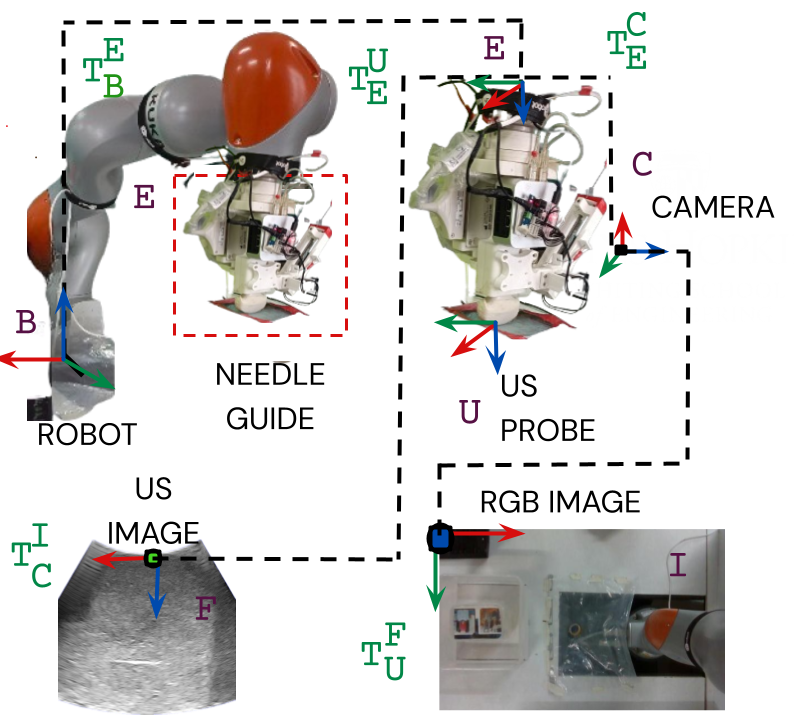}
    \caption{Showing coordinate system transformations between various components of the system.}
    \removeFigSpace
    \label{fig:sys_transforms}
\end{figure}

\subsection{Hardware Setup}
\label{sec:system_overview}

The presented system comprises the following components: (1) a 7-degree-of-freedom (DOF) robotic arm (KUKA LBR IIWA7 R800) with 7 Kg payload capacity for precise trajectory execution, (2) a Telemed (TELEMED MEDICAL SYSTEMS S.r.l., Milano, Italia) MicrUs EXT-1H with a convex array probe C5-2R60S-3 for real-time ultrasound imaging of the abdomen, (3) a Windows computer (Mini PC) equipped with a 12th Gen Alder Lake N100 processor (Quad-core, 1.7GHz-3.4GHz, 16GB RAM) and a Ubuntu Computer (host PC) with 32GB RAM for managing the sensor data and biopsy workflow respectively, (4) an RGB-D camera (Intel Realsense D435i) for the reconstruction of the 3D surface of the phantom, and (5) Intelligent Needle Guidance System (INGS) (1.6 kg weight) for autonomous needle placement. The INGS module can be attached to any robot with an end-effector holder and payload capacity of 2 kg. The INGS has a needle guide for autonomous needle insertion. The whole setup can be seen in Fig. \ref{fig:exp_setup_manqn} The open-source KST-Kuka-Sunrise Toolbox \cite{Safeea2019} and ROS (Robot Operating System) are used to control the KUKA LBR IIWA R800/R820). 

\begin{figure}[t!]
    \centering
    \includegraphics[width=\columnwidth]{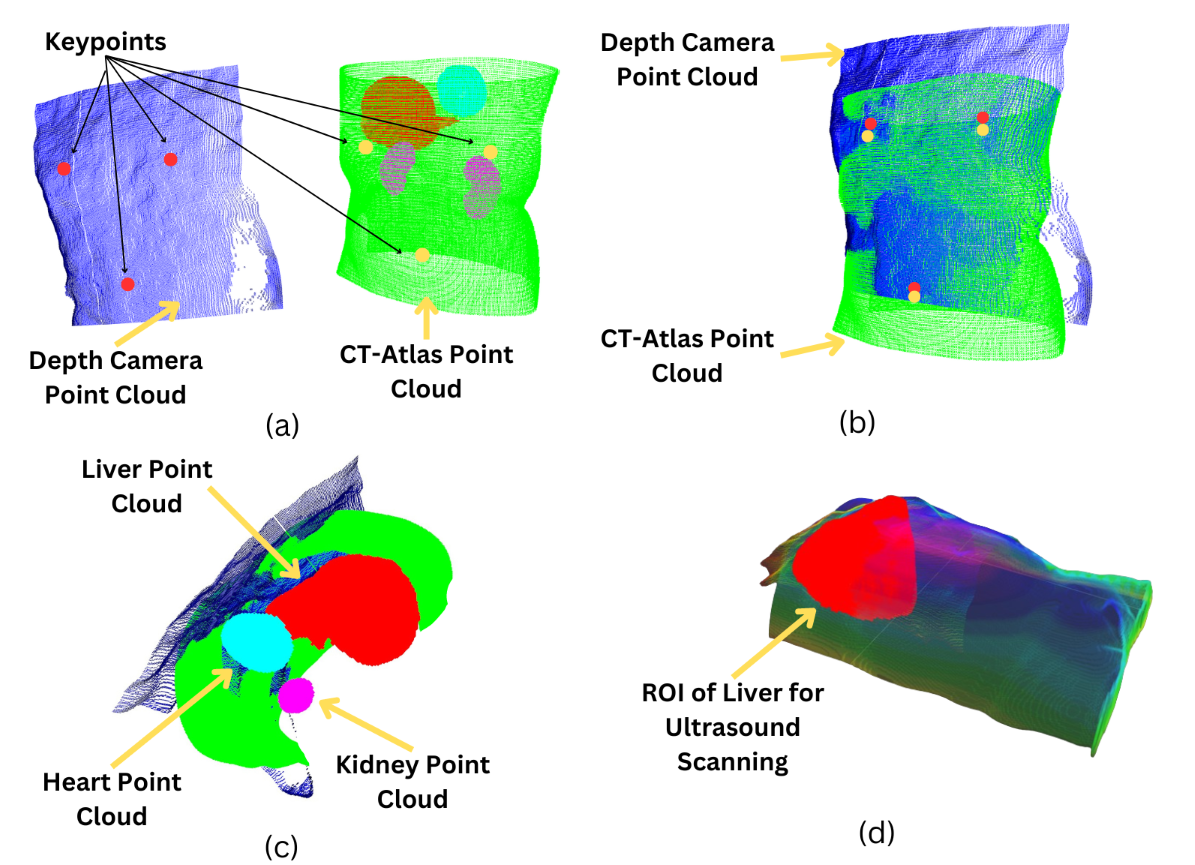}
    \caption{ATIR: Atlas-based Transformation and Iterative Registration: (a) Initial point clouds with keypoints
(b) Point clouds aligned with the initial transform generated by keypoint matching
(c) Point clouds aligned after refining the initial transform using ICP and corresponding 3D organs
(d) Corresponding ROI for Ultrasound scanning on the surface of the subject's point cloud 
*Only the liver is visualized, as scanning was limited to this organ for experiments}
\removeFigSpace
    \label{fig:organs_localization}
\end{figure}

The World Coordinate System (WCS) is defined at the robot's base. The 3D camera (C) frame $T^{C}_{E}$ and US probe (U) frame $T^{U}_{E}$ are defined with respect to the robot’s end effector frame  $E$ based on the CAD model. 2D image $I$ is converted to the point cloud (by pyrealsense library) using a camera matrix (\(T^{I}_{C}\)). The relationship between these transformation frames is shown in Fig. \ref{fig:sys_transforms}.

The RGB image (I) can be represented in WCS using the following equation:
\begin{equation}    
 T^{I}_{B} =  (T^{E}_{B}) \cdot (T^{C}_{E}) \cdot  (T^{I}_{C})
\end{equation}

Where \((T^{I}_{C})\) is provided by Intel RealSense, \((T^{C}_{E})\) is a fixed translation matrix described using CAD model, and \((T^{E}_{B})\) is calculated using KDL, an inverse kinematic solver.

Similarly, the 3D voxels in WCS for the 2D US image pixels are calculated using the following equation:
\begin{equation}    
T^{F}_{B} = (T^{E}_{B}) \cdot (T^{U}_{E}) \cdot (T^{F}_{U})
\end{equation}

Where \((T^{U}_{E})\) is a fixed transformation between the robot end effector and ultrasound probe derived from the CAD model. The resolution of the ultrasound image is 512 x 512. Physical dimensions of every pixel in US image are produced using the pixel spacing $[Sx, Sy] = [0.2, 0.2]~mm$, provided by Telemed Ultrasound Software Development Kit (SDK) and open-source PlusToolkit \cite{Lasso2014a}. Using the pixel spacing and transformation $T^{F}_{B}$, the 2D coordinates of pixels in the US image $(x, y)$ are converted to the 3D voxel coordinates defined as $(x_{F}, y_{F}, z_{F})$ using the following equations:


\begin{align*}
x_{F} &= x_{EE} + (256 - x) \cdot S_x \\
y_{F} &= y_{EE} \\
z_{F} &= z_{EE} + y \cdot S_y
\end{align*}

\((x_{F}, y_{F}, z_{F})\) represents the 3D voxel that lies in the US coordinate system. This point can be represented in WCS using:
\begin{equation}    
(x_{B}, y_{B}, z_{B}) = (T^{F}_{B}) \cdot (x_{F}, y_{F}, z_{F})
\end{equation}

Using this 3D voxel information, multiple image frames are converted into 3D US slices and stacked to produce the 3D reconstruction of the phantom. This 3D reconstruction is used further for lesion localization and targeting.


\subsection{ATIR: Atlas-based Transformation and Iterative Registration}
\label{sec:perception_algo}

ATIR estimates the position of target anatomical organs within the patient’s body. A generic CT-Atlas with segmented organs is prepared and registered to the patient, creating a detailed 3D anatomical map inside the anthropomorphic phantom. Fig. \ref{fig:organs_localization} shows the different stages of the registration procedure.

The CT-Atlas organs and body structures are generated by segmenting from a CT scan using 3D Slicer software. The abdomen CT data was segmented using the TotalSegmentor plugin \cite{Wasserthal_2023}. Separate point clouds are created for each organ and the abdomen skin surface point cloud. For the registration, DensePose \cite{guler2018densepose} has been used to filter the RGB-D point cloud such that only the subject's abdomen point cloud is retained. 

An accurate transformation matrix is estimated before applying ICP (Iterative Closest Point) to minimize axis mismatches between point clouds. Point clouds are typically initialized randomly, but using an estimated transformation matrix reduces randomness, accelerating the convergence process. The initial matrix is derived by aligning key points (left chest, right chest, navel) from the CT-Atlas with corresponding points on the anthropomorphic phantom, detected via an RGB-D camera, as shown in Figures \ref{fig:organs_localization} and \ref{fig:exp_setup_manqn}. 

Finally, after ICP (Iterative Closest Point) is applied, A convex hull is built around 3D points for the specific organ to be scanned. Points from the live abdomen point cloud within the convex hull are identified as part of the ROI, highlighting the region to be scanned on the human body for that particular organ.


\subsection{Robot Path Planning and Execution}
\label{sec:planning}

The path planning algorithm autonomously generates a sequence of robot poses on the phantom surface for ultrasound image acquisition. This is used to image and reconstruct the underlying volume beneath this area.

\subsubsection{Path Planning Algorithm and 3D Reconstruction}

The ROI generated on the patient body surface by the perception module as described in Section \ref{sec:perception_algo} is used to generate robot trajectory for ultrasound image acquisition. Point Cloud Library \cite{Rusu_ICRA2011_PCL} is used to compute normal \(\vec{z}_{\text{local}}\) of the local surface around the point \(\rho_{ij}\). The unit normal vector \( \vec{z}_{\text{local}} \) for each point  $\rho_{ij}$ can be used to calculate the orientation $\mathcal{O}_{ij}$ as follows :


\begin{equation}
\begin{aligned}
& \quad \vec{x}_{\text{local}} = \frac{\vec{z}_{\text{local}} \times \vec{y}_{\text{global}}}{\left\|\vec{z}_{\text{local}} \times \vec{y}_{\text{global}}\right\|} \\
& \quad \vec{y}_{\text{local}} = \frac{\vec{z}_{\text{local}} \times \vec{x}_{\text{local}}}{\left\|\vec{z}_{\text{local}} \times \vec{x}_{\text{local}}\right\|} \\
& \quad \mathcal{O}_{ij} = \begin{bmatrix} \vec{x}_{\text{local}} & \vec{y}_{\text{local}} & \vec{z}_{\text{local}} \end{bmatrix}
\end{aligned}
\end{equation}

where \( \vec{y}_{\text{global}} \) is the unit vector of the robot base in WCS (World Coordinate System). The unit vectors \( \vec{x}_{\text{local}} \), \( \vec{y}_{\text{local}} \), and \( \vec{z}_{\text{local}} \) at point $\rho_{ij}$ are used to form a rotation matrix $\mathcal{O}_{ij}$ of the robot's end effector at point $\rho_{ij}$. The complete trajectory that contains a set of robot’s end effector poses can be represented as:

\begin{equation}    
\mathcal{T} = \left\{ \rho_{ij} \{ (x_{ij}, y_{ij}, z_{ij})  ,\mathcal{O}_{ij} \}, | \, 1 \leq i \leq n, \, 1 \leq j \leq m \right\}.
\end{equation}

Moveit \cite{coleman2014reducing} is used to execute the above-generated task space trajectory. Moveit generates a joint space trajectory corresponding to the task space trajectory using KDL as an inverse kinematic solver; the generated joint space trajectory is executed by the robot using the KST-Kuka-Sunrise Toolbox \cite{Safeea2019}. 
Two different methods are used for the trajectory following, position and impedance control. The position control method moves the end effector to scan through a set of points without using the end effector forces. While, impedance control regulates a robot’s interaction with its environment by dynamically adjusting its compliance, or how it responds to external forces. This control strategy mimics a virtual spring-mass-damper system, where the robot’s interaction forces are modulated based on three key parameters: stiffness, damping, and mass. By fine-tuning these parameters, impedance control allows the robot to maintain constant probe-to-surface contact and ensure safety. The force safety threshold for impedance control was set to 12 Nm; the robot will automatically stop if the forces go beyond this threshold.


The robot acquires 2D US images at each point in the planned trajectory. The robot moves to the next point and repeats the process until the entire trajectory is executed. A 3D volume is constructed by stacking these 2D US images. Each 2D image is segmented to find the target (as described in Section \ref{sec:target_segm}), and transformed into a 3D slice (as described in Section \ref{sec:system_overview}).





\subsection{Lesion Segmentation and 3D Localisation}
\label{sec:target_segm}

To identify targets within the reconstructed ultrasound volume, the incoming 2D US image is pre-processed to reduce noise. The Circle Hough Transform (CHT) detects high-contrast circular targets in the pre-processed image. However, speckle noise and high-intensity echo waves could result in false target detection. A circular-intensity-based rejection (CIR) algorithm is applied to the resultant circular targets to check radial reduction in lesion intensity and discard the false target detection by CHT, as shown in equation \ref{eq:CIR}.


\begin{equation}   
\label{eq:CIR}
I(r + 2\delta) < I(r + \delta) < I(r)
\end{equation}

Where I(x) represents the average intensity of the pixels in the circle of radius x around a detected target and r is the radius of the detected tumor. $\delta$ is a threshold value that can be tuned based on the echo noise. If a particular detected target satisfies this equation it is considered as a valid target tumor since there is a continuous intensity drop over circles of larger radii. All target pixels inside the circle with an intensity greater than \( I_0 \) are converted into 3D target regions \( \mathcal{T}^{t}_{B} \) as mentioned in Section \ref{sec:system_overview}

The produced 3D volume is input to the density-based clustering non-parametric algorithm (DBSCAN). Its parameters, \( \epsilon \) and \( \gamma \), determine the maximum distance between points for clustering and the minimum points needed for a cluster, respectively. These parameters are tuned based on target density and neighboring distances to ensure consistent clustering. 

\begin{figure}[t!]
    \centering
    \includegraphics[width=\columnwidth]{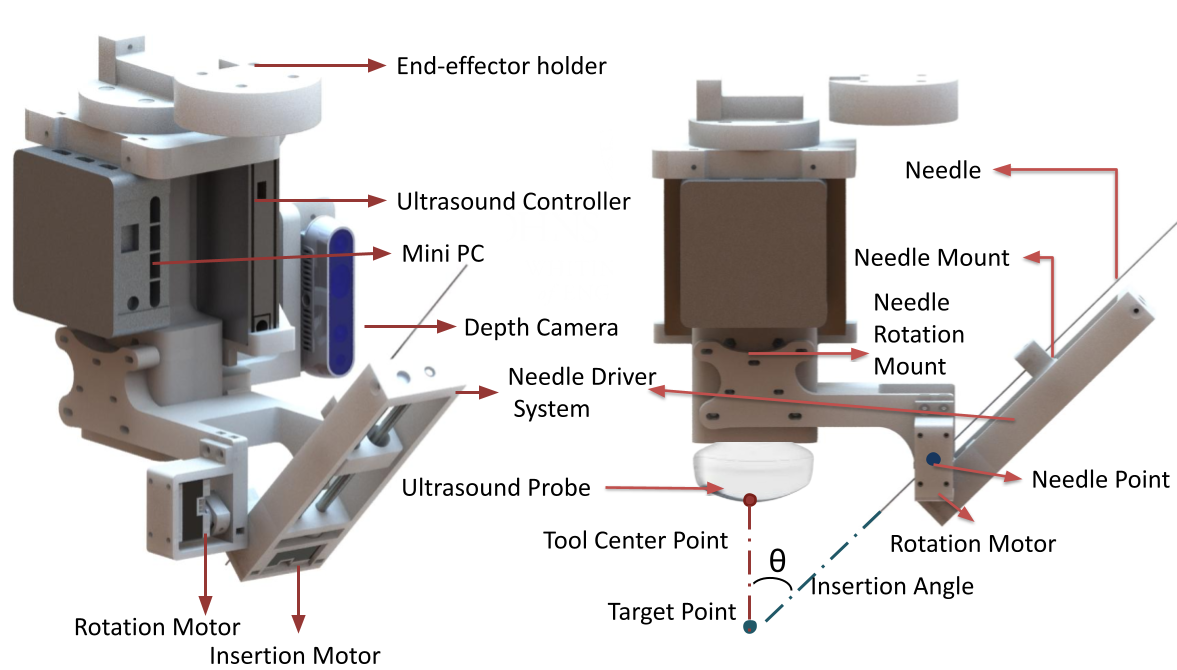}
    \caption{Needle driver assembly and insertion angle calculation for autonomous needle insertion.}
    \removeFigSpace
    \label{fig:INGS}
\end{figure}

\subsection{INGS: Intelligent Needle Guidance System}
\label{sec:needle_related}

The 2 DOF needle driver assembly shown in Fig. \ref{fig:INGS} consists of a needle driver module and a fixed mount. The standard biopsy needle can be attached to the needle guide. The needle guide can align the needle insertion axis to the desired needle insertion trajectory and insert up to a depth of 100 mm.

The insertion point is calculated using target centers identified by the DBSCAN algorithm. The insertion vector and the tool vector, denoted as \( \vec{v}_{\text{insertion}} \) and \( \vec{v}_{\text{tool}} \) respectively, are obtained as follows, $ \vec{v}_{\text{insertion}} and \vec{p}_{\text{target}} - \vec{p}_{\text{insertion}} $ and $ \vec{v}_{\text{tool}} and \vec{p}_{\text{target}} - \vec{p}_{\text{tool}} $


Here, \( \vec{p}_{\text{target}} \) represents the coordinates of the target point.
The insertion angle, denoted as \( \theta \), is determined using the dot product between the insertion vector and the tool vector and is calculated using the following equation: 

\begin{equation}
     \theta = \arccos\left( \frac{\vec{v}_{\text{insertion}} \cdot \vec{v}_{\text{tool}}}{\| \vec{v}_{\text{insertion}} \| \cdot \| \vec{v}_{\text{tool}} \|} \right) 
\end{equation}

The calculated angle is sent to the needle driver system to align the needle guide and perform needle insertion. The needle driver system uses two Dynamixel XC330-T288 motors for autonomous needle guide control, integrated with a Python GUI built using Tkinter and the Dynamixel library. The GUI continuously reads motor encoder positions, allowing real-time system monitoring. It also offers manual control for tasks such as testing, calibration, and homing. Homing ensures motors return to a zero reference point using mechanical stops. For US image to needle guide calibration, linear regression was used to establish the relationship between the needle guide angle and the pixel position of the needle entry point on the ultrasound image. The needle guide angle was varied from 30 to 40 degrees in 1-degree steps, and the corresponding pixel values were recorded. The relationship between the angle \(\theta\) and pixel position \(y\) is expressed as $\theta = \beta_0 y + \beta_1$

where \(\beta_0 = 0.03682\) and \(\beta_1 = 47.3703\). This allows the calculation of the needle guide angle based on the pixel position of the needle. Post-calibration, the average error decreased from 0.69 ± 0.37 mm to 0.35 ± 0.24 mm.


\begin{figure}[t!]
    \centering
    \includegraphics[width=\columnwidth]{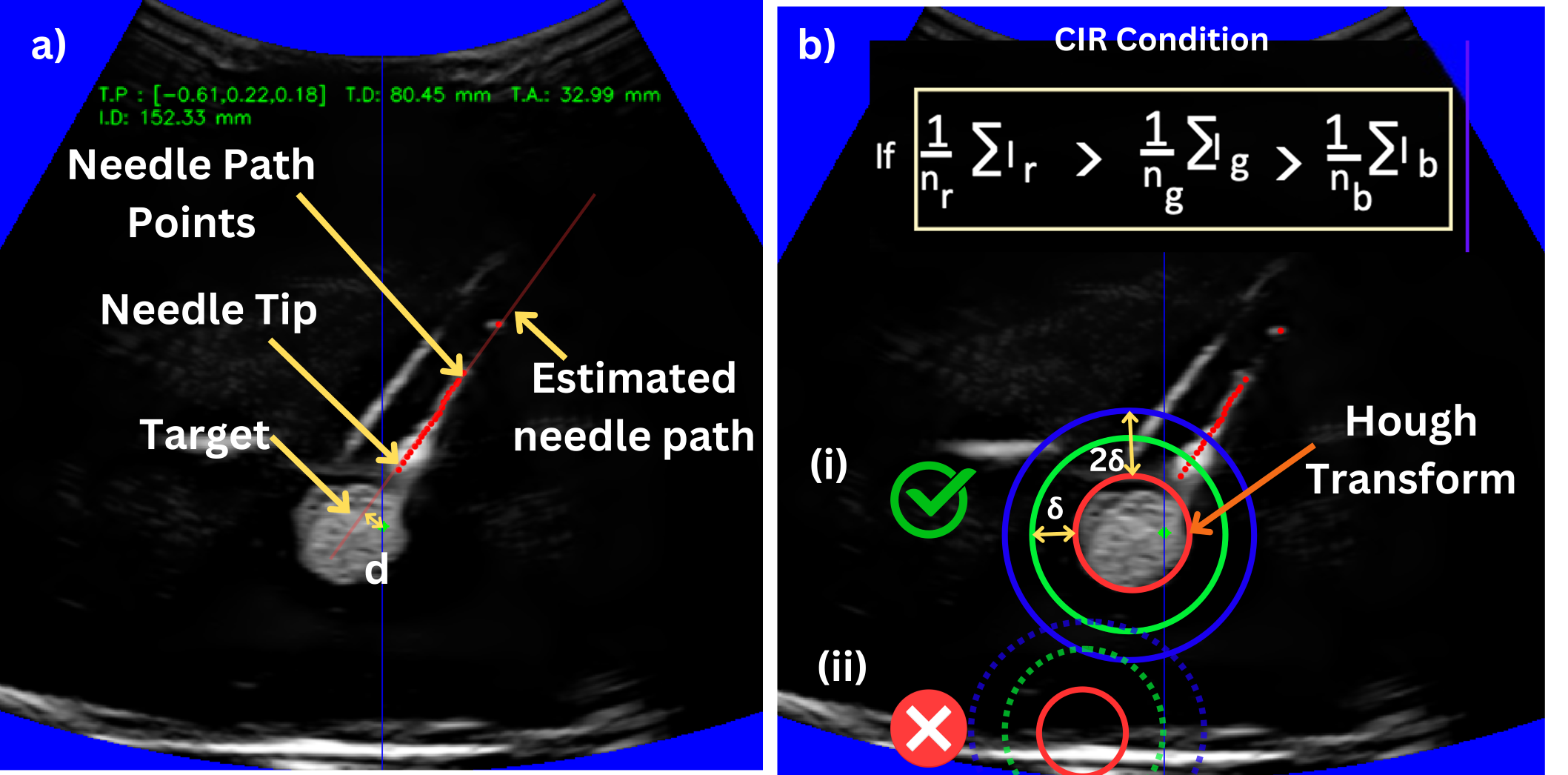}
    \caption{a) shows the needle targeting error calculation in which user manually selects point on the visible needle path in US image. b) shows the CIR condition where case (i) shows the valid target as the average intensity of CIR output is higher than the surrounding region, and case (ii) shows an invalid target as the average intensity of CIR output is comparable to the neighborhood region. }
    \removeFigSpace
    \label{fig:error_calculation}
\end{figure}


\begin{table*}[htbp!]
\centering
\caption{Comparison of Mean Targeting Error, Mean Time, and Target Locations for Position and Impedance Control modes}
\begin{tabular}{lcccccc}
\toprule
\textbf{Target} & \textbf{Control Type} & \textbf{Target Point X} & \textbf{Target Point Y} & \textbf{Target Point Z} & \textbf{Mean Error (mm)} & \textbf{Mean Time (s)} \\
\midrule
Target 1 & Position  & -603.59 & 223.81 & 155.81 & 4.582 & 29.682 \\
Target 2 & Position  & -622.97 & 222.04 & 175.03 & 3.168 & 12.641 \\
Target 3 & Position  & -576.68 & 226.54 & 168.54 & 2.163 & 23.826 \\
\midrule
\textbf{Overall} & Position & - & - & - & 3.304 & 22.049 \\
\midrule
Target 1 & Impedance & -581.15 & 225.76 & 166.80 & 7.530 & 32.979 \\
Target 2 & Impedance & -553.72 & 231.30 & 171.16 & 4.330 & 15.947 \\
Target 3 & Impedance & -604.97 & 222.67 & 180.23 & 5.365 & 31.218 \\
\midrule
\textbf{Overall} & Impedance & - & - & - & 5.741 & 26.715 \\
\bottomrule
\end{tabular}
\label{tab:results}
\end{table*}

\section{Experiments and Results}
\label{sec:exp}

The targeting accuracy of the presented autonomous robotic biopsy system is evaluated using an anthropomorphic phantom. Three foam targets ranging from 15 to 20 mm in size were embedded in an agar-water-based phantom. This agar-water mixture with the ratio of 2 g agar to every 100 ml of water
was boiled for 15 minutes. This phantom was fixed at an approximate liver location in the anthropomorphic phantom. 


The autonomous targeting procedure begins with the homing of the needle guide; then, the robot moves to a fixed camera position to capture the surgical scene using the RGBD camera. The captured RGB image is used to identify the abdomen region, static key points, and the rectangular safety workspace. ATIR (Sec. \ref{sec:perception_algo}) processes the detected key points and the abdomen region to detect the organ of interest from the abdomen point cloud. Following this, a path is generated, and the robot executes the trajectory in a point-to-point manner, skipping any points outside the safety workspace. During the execution, the system captures an ultrasound (US) image at each point on the trajectory, identifies potential targets, and converts the segmented image into a 3D US slice in the robot’s base coordinate system. These slices are then concatenated to form a US volume, which is processed using a clustering algorithm to localize all targets and their centroid. Once targets are localized, the robot plans and executes a trajectory to reach each target's centroid, calculating the insertion angle and depth for precise needle insertion and retrieval. 

Experiments were performed with two robot control modes (1) only position-controlled trajectory execution and (2) impedance-controlled trajectory execution. For each of these control modes, experiments were repeated ten times, resulting in a total of 30 targeting attempts for each of the control modes.

\subsection{Accuracy Assessment}
\label{sec:accuracy}

The positional accuracy of autonomous needle targeting was assessed by manually selecting the needle pixels on ultrasound (US) images saved after each needle insertion, as shown in Figure \ref{fig:error_calculation}. These selected needle pixel positions on the US image were then converted to 3D needle points using the saved robot pose. To avoid user bias, the needle trajectory was manually assessed three times. Principal Component Analysis (PCA) was used to fit a line to the 3D needle points. The targeting error is calculated as the minimum distance between the 3D line and the target point.




\subsection{Results}
As shown in Table \ref{tab:results}, the mean positional accuracy of the autonomous percutaneous system is $3.304~\pm~2.70~mm$ in position-controlled mode and $5.74~\pm~2.70~mm$ in impedance-controlled mode. The increase in targeting error in impedance-controlled mode can be attributed to force compliance adapted to maintain probe-to-surface contact, resulting in adjustments in the probe's position while performing US scan and needle placement. The total time to reach the target in position-controlled mode was 22.04 seconds, while impedance-controlled mode took 26.71 seconds. Although the needle targeting time should be comparable, the difference in time is due to fewer needle insertions with effective needle visibility in position-controlled mode. A lower error and lower sampling time for position control can be attributed to the fact that the position control system missed targets during the scanning procedure in a few attempts. Each target was supposed to be sampled 10 times each (a total of 30 attempts for 3 targets); however, in position control mode, only 18 attempts were successful, while in the impedance control model, all 30 attempts were successful. 
This discrepancy is due to uneven probe-to-surface contact on the phantom, caused by deformation from the robot-held US probe and slight mechanical misalignment of the needle guide, which tilted the needle off the US imaging plane in the case of position control. However, impedance-controlled percutaneous operation maintained consistent and safe surface contact throughout the procedure. This is a clear indication of the requirement of an impedance-controlled system for clinical translation. As corroborated in \cite{odegaard1992effects} uneven force on the surface by the ultrasound probe can cause several errors. These results highlight the potential of an autonomous impedance-controlled percutaneous system for safe and effective liver biopsy with tumor size larger than 5 mm.

\section{Conclusion and Discussion}

This paper presented a fully autonomous robotic system for ultrasound-guided percutaneous interventions. The system operates through a multi-step process designed to achieve full autonomy. Firstly,  target organs are localized using CT-point cloud registration using the ATIR algorithm, followed by generating a region of interest for the organ on the subject's skin. The system then plans and executes a path for ultrasound scanning. Two robot control modes have been evaluated, position and impedance control for trajectory execution. During this scanning process, 2D ultrasound images are acquired, with tumor detection facilitated by the CIR algorithm. These 2D images are stacked to produce a 3D volumetric representation of the scanned organ and lesions. Finally, the calibrated needle guidance system is used to perform needle insertion. Using the impedance control mode, the experiments showed a targeting accuracy of $5.74 \pm 2.7$ mm, lesion detection and targeting success rate of 100\%, and a mean procedure time of 26.7 seconds. These results demonstrate the potential for autonomous ultrasound-guided interventions. With the achieved accuracy, lesion detection rate, and successful needle placements, the presented system shows promise for achieving full autonomy in performing US-guided percutaneous interventions. Future work will focus on \textit{ex vivo} studies with nonhomogeneous phantoms and organ motion compensation.


\bibliographystyle{IEEEtran}
\bibliography{references}

\end{document}